\documentclass[conference, a4paper]{IEEEtran}
\IEEEoverridecommandlockouts
\usepackage{cite}
\usepackage{amsmath,amssymb,amsfonts}
\usepackage{algorithmic}
\usepackage{graphicx}
\usepackage{textcomp}
\usepackage{xcolor}
\usepackage{tikz}
\usepackage{comment}

\usepackage[hidelinks]{hyperref}
\ifCLASSOPTIONcompsoc \usepackage[caption=false,font=normalsize,labelfon t=sf,textfont=sf]{subfig} \else \usepackage[caption=false,font=footnotesize]{subfig} \fi
\usetikzlibrary{positioning,fit}

\def\BibTeX{{\rm B\kern-.05em{\sc i\kern-.025em b}\kern-.08em
    T\kern-.1667em\lower.7ex\hbox{E}\kern-.125emX}}

\usetikzlibrary{arrows,shapes}
\tikzset{
	vertex/.style = {
		circle,
		fill            = black,
		outer sep = 2pt,
		inner sep = 1pt,
	},
	block/.style    = {draw, thick, rectangle, minimum height = 1em,
						minimum width = 1.5em},
	sum/.style      = {draw, circle, node distance = 1.5cm}, 
	neuron/.style   = {draw, circle, node distance = 1cm}, 
	input/.style    = {coordinate}, 
	output/.style   = {coordinate} 
}

\begin{document}

\title{Capturing and Explaining Trajectory Singularities\\using Composite Signal Neural Networks}

\author{
	\IEEEauthorblockN{Hippolyte Dubois\IEEEauthorrefmark{1}, Patrick Le Callet\IEEEauthorrefmark{1}, Michael Hornberger\IEEEauthorrefmark{2}, Hugo J. Spiers\IEEEauthorrefmark{3}, Antoine Coutrot\IEEEauthorrefmark{1}}
	\IEEEauthorblockA{\IEEEauthorrefmark{1}Université de Nantes, CNRS, LS2N, F-44000 Nantes, France
    \\\{hippolyte.dubois, patrick.lecallet,  antoine.coutrot\}@univ-nantes.fr
	}
	\IEEEauthorblockA{\IEEEauthorrefmark{2}
	Norwich Medical School, University of East Anglia, Norwich, United Kingdom
	}
	\IEEEauthorblockA{\IEEEauthorrefmark{3}
	Institute of Behavioural Neuroscience, Department of Experimental Psychology,
	\\Division of Psychology and Language Sciences,\\
	University College London, London, United Kingdom
	}
}
\maketitle

\begin{abstract}
	Spatial trajectories are ubiquitous and complex signals. Their analysis is crucial in many research fields, from urban planning to neuroscience. Several approaches have been proposed to cluster trajectories. They rely on hand-crafted features, which struggle to capture the spatio-temporal complexity of the signal, or on Artificial Neural Networks (\textbf{ANNs}) which can be more efficient but less interpretable. In this paper we present a novel ANN architecture designed to capture the spatio-temporal patterns characteristic of a set of trajectories, while taking into account the demographics of the navigators. Hence, our model extracts markers linked to both behaviour and demographics. We propose a composite signal analyser (\textbf{CompSNN}) combining three simple ANN modules. Each of these modules uses different signal representations of the trajectory while remaining interpretable. Our CompSNN performs significantly better than its modules taken in isolation and allows to visualise which parts of the signal were most useful to discriminate the trajectories.
\end{abstract}


\section{Introduction}

	Understanding and modelling the basic laws governing human spatial navigation is crucial is many fields such as urban planning\cite{tang_efficient_2017}, traffic forecasting\cite{besse_review_2016}, activity understanding\cite{alahi_social_2016},  ecology\cite{streitz_understanding_2018}, behavioural and clinical neuroscience \cite{Coughlan2018hi}, see \cite{DamasceneMazimpaka2016eo} for a review. The rapid spread of GPS devices, sensor network, satellite and wireless communication technology, enables the tracking of all kinds of moving objects all over the world. This results in an increasing number of moving object trajectory data to be collected and stored in large-scale databases \cite{tong2020spatial}. 
	Spatial trajectories are rich signals that are not straightforward to handle, notably because they are determined by a complex interaction between the shape of the environment (walls, obstacles: bottom-up determinants) and the navigator’s decisions (task at hand, cognitive state: top-down determinants). 
	To mine these complex spatial patterns, numerous approaches have been proposed in the literature. The most popular approach is trajectory clustering, which aims at discovering groups of similar trajectories in an unsupervised, semi-supervised, or supervised manner. Trajectory clustering typically quantifies trajectories with a set of features (e.g. coordinates, angle, speed, curvature…) and then apply clustering algorithms such as \textbf{K-means} or Density-Based Spatial Clustering of Applications with Noise (\textbf{DBSCAN}) \cite{Yuan2017bk}. Another approach is to use trajectory similarity measures such as dynamic time warping (\textbf{DTW}), edit distance on real sequence (\textbf{EDR}), longest common subsequences (\textbf{LCSS}) and then apply a clustering algorithm in the space defined by these metrics \cite{bian_survey_2018}.

	These classical data mining techniques have the advantage of being highly interpretable, but they are quite limited for several reasons. First, trajectories intrinsically have spatial and temporal properties, which conventional approaches often can not handle at the same time. Second, trajectories are strongly self-correlated (position at time $t$ is highly dependent on position at time $t-1$). This violates the assumption of independence of samples, on which many conventional clustering approaches depend. To overcome these limitations, artificial neural networks (\textbf{ANN}) have been proposed with success, as they can learn efficient representations directly from the raw data, without the need to hand-craft the features \cite{Wang2019vq}. ANN can also jointly capture spatial patterns (e.g. with convolutional neural networks, \textbf{CNN}) and temporal patterns (e.g. with recurrent neural networks, \textbf{RNN}). The main limitations of ANNs for trajectory analysis is that while being efficient in a certain context, they poorly generalize to other situations. Furthermore, while ANN explainability has recently been in the spotlight in some research fields (particularly in medecine \cite{holzinger2019causability}), this has less been the case in trajectory analysis, where they mostly remain black-boxes.

	\section{Contributions}
    In this study we propose a method to capture the spatio-temporal patterns characteristic of the trajectories followed by a given population navigating in a constrained environment. We aim to associate these patterns to the demographics of the navigators, to extract clusters homogeneous in both \textit{behaviour} and \textit{demographics}.
	To extract meaningful and interpretable trajectory markers, we propose a composite signal analyser (\textbf{CompSNN}). We hypothesize that, by  combining simple models capturing different aspects of the data, we can obtain good analysis performance, while maintaining a high level of explainability.
	Our analyser uses three dependencies between the samples of our data: 
	\begin{itemize}
		\item Temporal dependency using time-series representation, with a CNN
		\item Spatial dependency using graph representation, with a Multi Layer Perceptron (\textbf{MLP})
		\item Spatial dependency using graph spectrum representation, with a Graph convolutional Neural Network (\textbf{GCNN})
	\end{itemize}
	
	The output of those networks is then concatenated, and fed into a MLP, which combines temporal and spatial signal analysis, hence the Composite Signal. This MLP is trained to predict the demographic from the behaviour. Our code is available on \url{https://gitlab.univ-nantes.fr/E173825Q/compsnn-eusipco2020}.

\section{Architecture of the CompSNN model}
    \begin{figure*}[t]
        \centering
        \includegraphics[width=.7\linewidth]{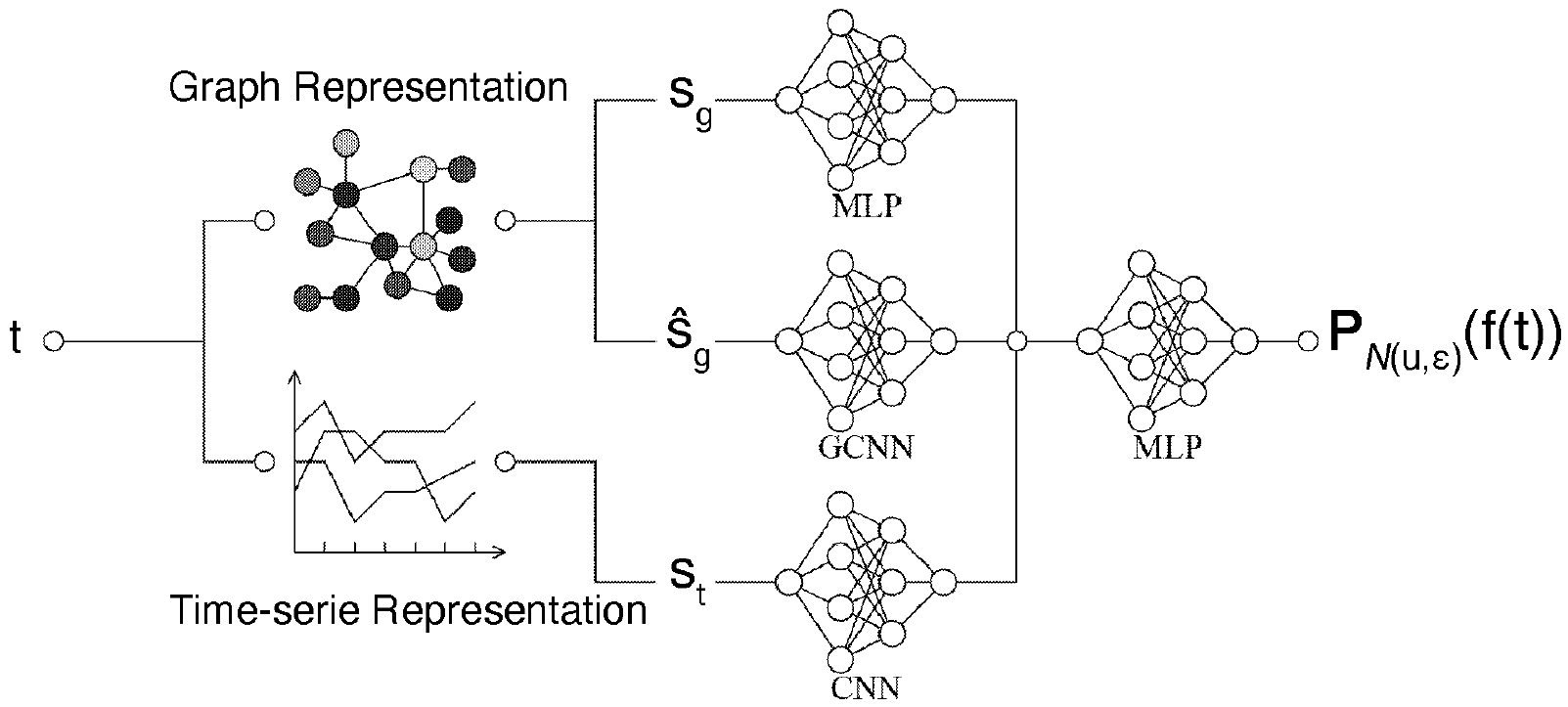}
        \caption{Architecture of the CompSNN model. Trajectory \textbf{t} is represented by 1- a signal on a graph $s_g$, 2- its Graph Fourier Transform $\hat{s}_g$ and 3- time-series of trajectory features. These representations are respectively fed to a MLP, a GCNN and a CNN. The outputs of these modules are then aggregated by a MLP trained to predict the demographics from the signal extracted from the behaviour.}
        \label{fig:archi_compsnn }
	\end{figure*}
\begin{figure*}[t]
	\centering
	\includegraphics[width=.4\linewidth]{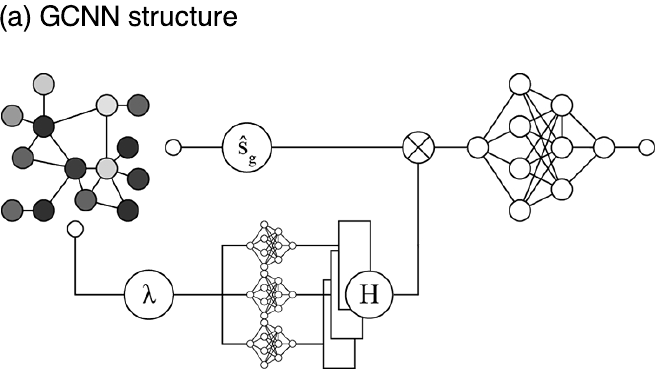}
\includegraphics[width=.3\linewidth]{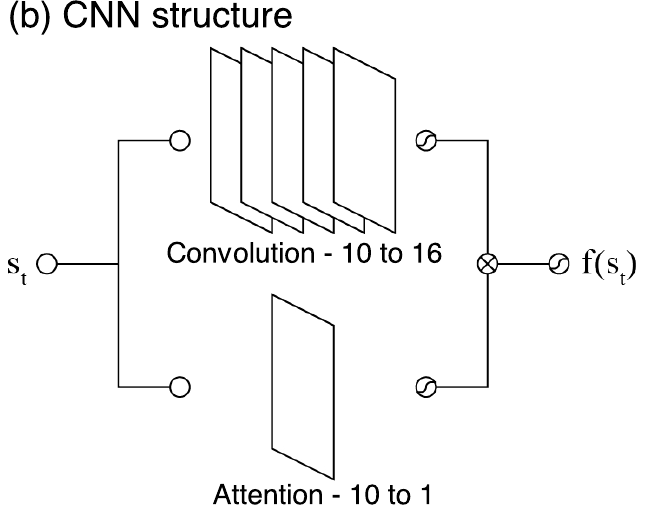}
	\caption{Structures of the GCNN and CNN components. (a) GCNN structure, with $H$ the filterbank, and $\lambda$ the eigenvalues of $\mathcal{G}$. (b) CNN structure. The output of each layer is constrained by a Sigmoid function. }
	\label{fig:modules}
\end{figure*}
    The CompSNN is an aggregate of three different ANNs: a MLP, a GCNN, and a CNN. In this section, we describe each of these components.

\subsection{Components}

\paragraph{MLP}
    The first component is a MLP taking the signal defined on the graph $\mathcal{G}(\mathbf{N},\mathbf{E})$ nodes as input. It is agnostic of the graph's structure, so it's important that the graph modeling the data is designed to capture the singularity of the signal. We fully describe the graph layout section (see \figurename~\ref{definegraph}). 
    
	The MLP is composed of a first linear layer, with $\mathbf{|N|*8}$ inputs, $32$ hidden nodes, and $16$ outputs, with a ReLU non-linearity activation in the middle. It processes signal defined on the nodes of the graph, each node holding values in $\mathbb{R}^8$, namely:
	\begin{itemize}
		\item the mean of the speed $s$,
		\item the mean of the acceleration $\nabla s$,
		\item the mean of the direction $\theta=atan2(x,y)$,
        \item the mean of the curvature $\nabla \theta$,
        \item the mean of the entropy $\mathbf{s}(\nabla \theta)$,
        \item the mean of the variance $\sigma^2(\nabla \theta)$,
        \item if the navigator exited the node then came back,
        \item the number of time it was on the node.
	\end{itemize}
	
\paragraph{GraphCNN}
    The second component is a Graph convolutional Neural Network (see \figurename~\ref{fig:modules}a), that learns a set of $j$ polynomial filters $H$. Each of those filters is the output of a MLP which takes the eigenvalues of the graph $\mathcal{G}$ as input \cite{shuman_emerging_2013}. The input signal, that is the number of time the navigator was on each node, is then processed as follows:
	
	\begin{equation}
		\begin{aligned}
			h_k &= MLP_{h_k}(\lambda)\\
			g(h) &= \sum_{k=0}^K h_k \Lambda^k\\
			f_h(s) &= g(h) \times \hat{s}\\
			f(s) &= \sum_{h \in H}\oplus f_h(s)
		\end{aligned}	
	\end{equation}

    with $\times$ the piecewise product of two vectors, $\Lambda = diag(\lambda)$, $\lambda_i$ the i-th eigenvalue of the graph's Laplacian, and $\sum \oplus$ the concatenation of several vectors.

	As product in spectral domain is equivalent to convolution in spatial domain, this is essentially a convolutional layer. One could argue that this is not rigorously a GCNN as the filters are not localized, but, because we do not go back to the spatial domain and we do not stack convolutional layers, we do not think it is necessary to implement localization, as it makes the model a lot more computationally expensive.

    The output of the convolutional layer is then fed into a linear layer with $k\mathbf{|N|}$ inputs and $16$ outputs.

\paragraph{CNN}

    The third component is a straight-forward 1D CNN (see \figurename~\ref{fig:modules}b), with a convolutional layers of 16 features, as well as an attention layer with a 1-dimensional output, followed up with a linear layer with 16 outputs. The output of the feature-convolutional layer is multiplied by the weight outputed by the attention-convolutional layer, both of those layers providing knowledge about the data structure. Because there is only one feature-extracting layer, the model remains explainable.

\paragraph{Aggregator}
	All those components feed into a MLP with $16+16+16=48$ inputs and $|u|$ outputs.

\subsection{Loss Function}
    The aggregator's output is evaluated by computing the log of the probability density function (\textbf{PDF}) of the distribution $\mathcal{N}(u,\epsilon)$. The SGD optimizer then tries to minimize the opposite of the log of the normalized $PDF(x)$.

	\begin{equation}
		loss(x,u) = - \log\left(PDF_{\mathcal{N}(u,\epsilon)}(x) - PDF_{\mathcal{N}(u,\epsilon)}(u)\right)
	\end{equation}

	$\mathcal{N}(u,\epsilon)$ is the gaussian distribution centered around the N-dimensional point $u$ sampled from the demographic space, with $\epsilon$ the standard deviation of the distribution.
	
	We normalize the PDF because in order to compare models with different $\epsilon$ parameters, and to allow the model to know when it is right, as the maximum of the PDF is not $1$. This loss function is used because, as the data is noisy, it is important to include a way to mitigate that noise when training, for example by providing the target-space structure by setting the $\epsilon$ parameter to be the variance of the demographic domain.

\section{Trajectory dataset}
	We use trajectories recorded in the Sea Hero Quest (\textbf{SHQ}) project \cite{coutrot_global_2018}. SHQ is a mobile video game designed to quantify the spatial ability of players. It collected data from over 4 million people worldwide. Here we use trajectories recorded during "checkpoint levels", where players were initially presented with a map indicating the starting point and the location of several checkpoints to find in a set order. Then the map disappeared and players had to navigate their way to the checkpoints. Their trajectories $T$ was recorded at 2 Hz, along with demographic information $U$ known to interact with spatial ability (age, gender, education, home environment, sleep duration, commute duration, handedness and self-assessment of spatial ability).
	The performance at SHQ has been shown to be highly correlated to the performance at a similar spatial task in the real world \cite{coutrot_virtual_2019}. Previous analyses of this dataset showed that spatial ability was indeed modulated by the player’s demographics such as age and gender \cite{coutrot_global_2018}. However these results were mostly based on the length of the trajectories, which can not fully capture the spatial strategies used by the different demographic groups. For instance, two trajectories can have the same length while having completely different shapes.
	
\begin{figure}[!ht]
		\centering
		\subfloat[]{\includegraphics[width=.3\linewidth]{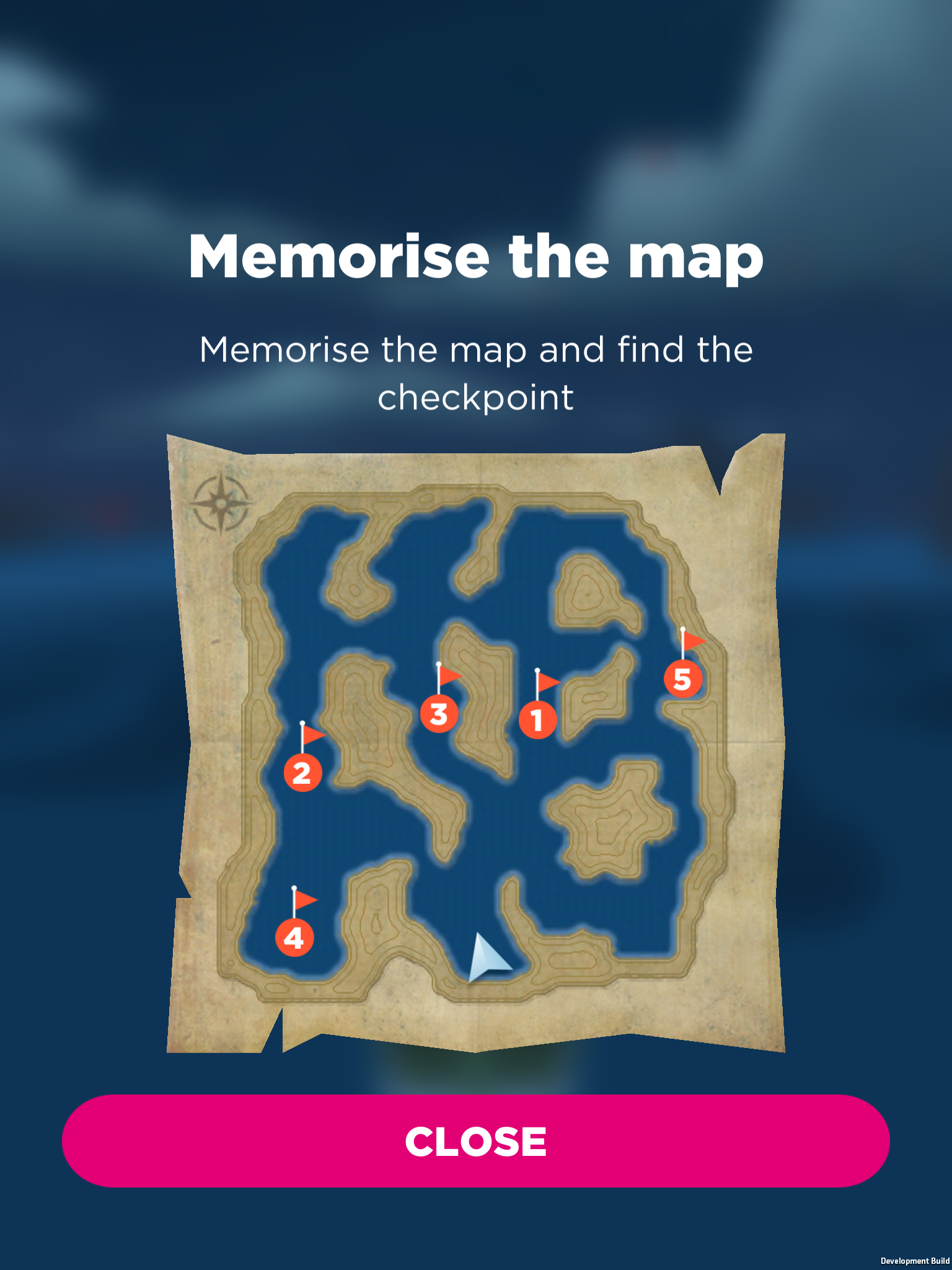}
		\label{fig_view}}
		\subfloat[]{\includegraphics[width=.3\linewidth]{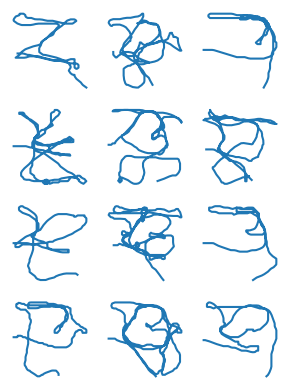}
		\label{fig_trajs}}
		\caption{The Sea Hero Quest Dataset - (a) Map of level 48. Players had to memorize the location of the 4 checkpoints and navigate to them in a set order. (b) Twelve random resulting trajectories (out of 39,289).}
		\label{fig_examples}
	\end{figure}

    Each trajectory $t$ of the set $T$ (see \figurename~\ref{fig_trajs}) is composed of a $x$ and a $y$ component, captured $N$ times, with $N$ being heterogeneous among $T$. Therefore we have $t \in \mathbb{R}^{2 \times N}$.
	
    We can increase the dimensionality of $t$ by computing at each timestamp $n$:
    \begin{itemize}
        \item the speed $s$,
        \item the acceleration $\nabla s$,
        \item the $x$ derivative $\nabla x$,
        \item the $y$ derivative $\nabla y$,
        \item the direction $\theta=atan2(x,y)$,
        \item the curvature $\nabla \theta$,
        \item as well as the local entropy $\mathbf{s}(\nabla \theta)$
        \item and the local variance $\sigma^2(\nabla \theta)$.
    \end{itemize}

    We therefore have $t\in \mathbb{R}^{10\times N}$

    Each user is defined by its demographic vector $u \in [0,1]^8$. Categorical values are arbitrarily assigned values in the $[0,1]$ interval, and ordinal values are ordered from $0$ to $1$ with a regular spacing.
    
    The CompSNN defined earlier will try to learn a function $f(t|\lambda)$, with $\lambda$ the parameters of the model, st.

    \begin{equation}
        \max_{\lambda} \mathbb{P}_{\mathcal{N}(u,\epsilon)}(f(t|\lambda))        
    \end{equation}
    
	with $\epsilon$ a meta-parameter used to allow the network to fail. $\epsilon$ is here set to $\sigma(U)$.

\section{Defining the graph}
\label{definegraph}
	It is important that the structure of the graph used to quantize the input signal is fitted to our task. To capture the singularity of the signal, while maintaining the number of nodes low to avoid redundancy and reduce the complexity of the model, the graph is defined by segmenting the heatmap of the level, with the inverse density probability as a weight. That way, areas that are visited by most players are aggregated, and areas that are more significant to identifying variance, that is area that are visited only by the few, are less packed.

\begin{figure}[!ht]
		\centering
		\subfloat[Inverted distribution]{\includegraphics[width=.4\linewidth]{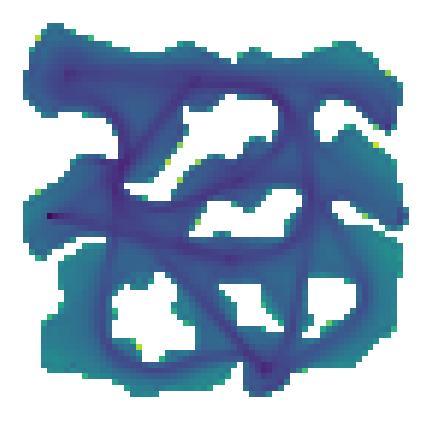}
		\label{fig_invdistmap}}
		\subfloat[Local maximas]{\includegraphics[width=.4\linewidth]{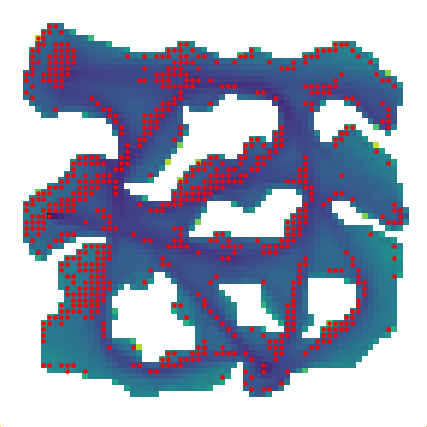}
		\label{fig_locmax}}
		\hfil
		\subfloat[Segmented map]{\includegraphics[width=.4\linewidth]{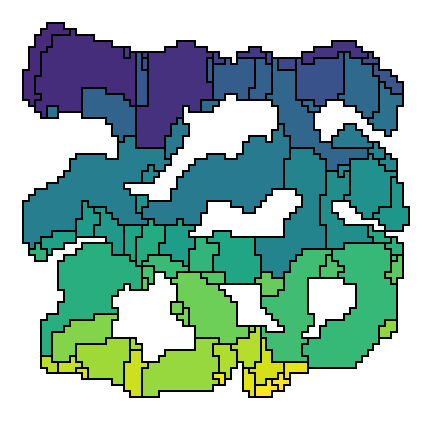}
		\label{fig_segmap}}
		\subfloat[Generated graph]{\includegraphics[width=.4\linewidth]{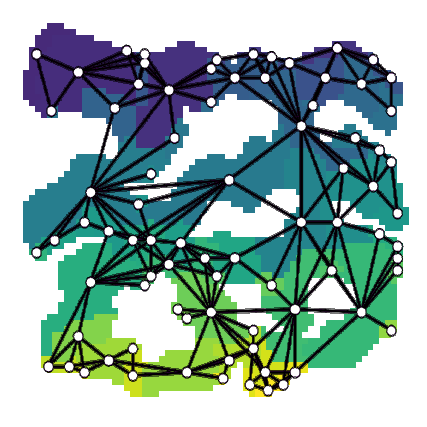}
		\label{fig_mapgraph}}
		\caption{Segmentation of the level 48 map using the watershed algorithm}
		\label{fig_watershed} 
	\end{figure}

	To do so, we use a watershed algorithm, with the local maximas (see \figurename~\ref{fig_locmax}) of the original probability density $\mathcal{D}_d$ in the $\{x,y\}$ plane as centers, with $d$ the level. The watershed algorithms then propagates from those centers using the inverse $\frac{1}{\mathcal{D}_d+1}$ density (see \figurename~\ref{fig_invdistmap}). Each segment of the map (see \figurename~\ref{fig_segmap}) is associated to a node of the graph $\mathcal{G}(\mathbf{N},\mathbf{E})$. The edges are computed from the trajectories, so that $\exists \mathbf{E}_{ij}$ if at least one player went from the area associated to $\mathbf{N}_i$ to the one associated to $\mathbf{N}_j$ (sees \figurename~\ref{fig_mapgraph}).
	
	The graph definition could be enhanced by using a higher dimensional distribution map, for exemple in $\{x,y,atan2(x,y)\}$. The trade-off is the increase in dimensionality of the graph signal as the number of nodes increases.

\section{Evaluation}
To demonstrate the added value of using three different modules to capture different aspects of the signal, we will compare the results of the components on their own to the ones of the composite model.

\subsection{Data}
	To run the experiment, we used a subset of the 39,289 trajectories sampled from SHQ level 48 with full demographic information. We used 1000 trajectories as it is enough to show how the model performs. We used trajectories from level 48 as it is a complex level with a challenging topology, likely to elicit a larger variance in behavior than simpler levels.
\subsection{Structure of the \textit{SingleNNs}}
	To compare the performance of the CompSNN to the individual modules, we train SingleNNs, which are models with a module (graph signal MLP, GCNN or CNN) feeding into a linear model with 16 inputs and $|u|$ outputs, evaluated with the same loss as the CompSNN. The weights are not shared between the CompSNN modules and the SingleNNs.

\subsection{Comparison}

\begin{figure}[!ht]
	\centering
	\subfloat[Mean loss per model over epochs.]{\includegraphics[width=.8\linewidth]{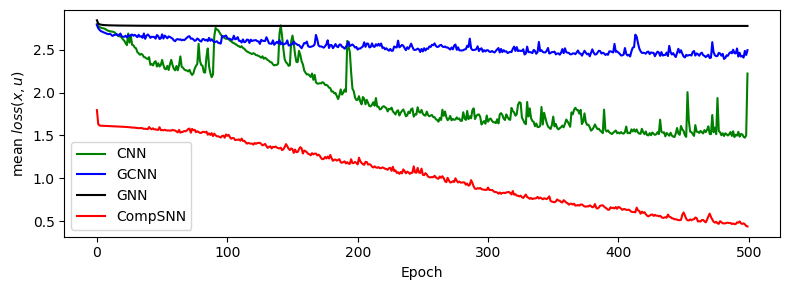}
	\label{fig_losseps}}\\
	\subfloat[Loss histogram at best epoch]{\includegraphics[width=.8\linewidth]{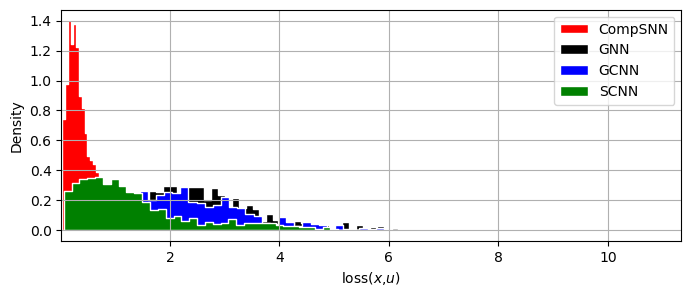}
	\label{fig_losshist}}\\
	\subfloat[Mean and $95\%$ Confidence Interval at the best epoch]{
		\small
		\begin{tabular}{l r r r r}\hline
		Model & CNN & GCNN & GNN & CompSNN \\\hline
		Mean & 1.47 & 2.39 & 2.78 & 0.44 \\
		CI & [1.40;1.54] & [2.32;2.46] & [2.71;2.84] & [0.42;0.45] \\\hline		
		\end{tabular}
	}
	\caption{Comparison of the scores for each model}
	\label{fig_hist}
\end{figure}

\begin{figure}[!ht]
	\centering
	\includegraphics[width=.6\linewidth]{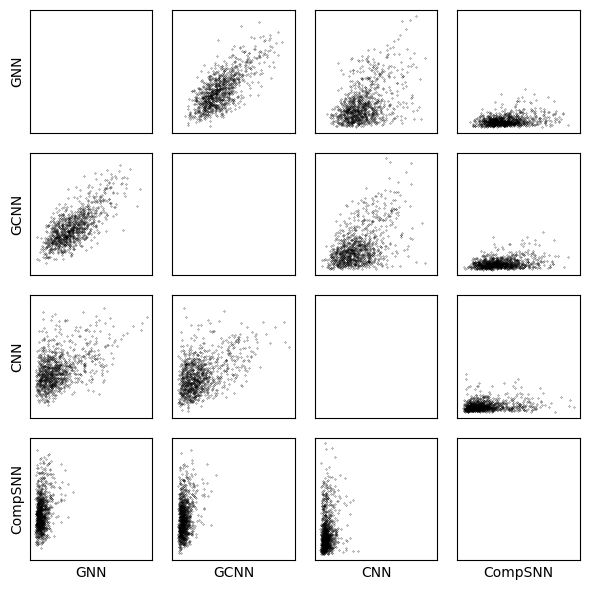}
	\caption{Correlations between CompSNN and SingleNNs scores.}
	\label{fig_corr}
\end{figure}

By comparing the models' performances, we can see, that, while keeping the same \textit{deepness}, the CompSNN performs significantly better (see \figurename~\ref{fig_hist}). By providing other representations of the signal, making it \textit{wider}, we can improve the efficiency of our system, while keeping it \textit{shorter}. As \figurename~\ref{fig_corr} shows us, there is a correlation between the score of samples across SingleNNs, even though the CNN performs better. This suggests that a sample's \textit{hardness} is not correlated to its signal representation, there is not a representation that would be more fitted for some samples but not others, but rather that the information can be found by aggregating different representations. 

\section{Where the information lies}
Because we keep the system short, we can use the learned weights to understand our data.

\begin{figure}[!ht]
	\centering
	\subfloat[Attention $a$]{\includegraphics[width=.4\linewidth]{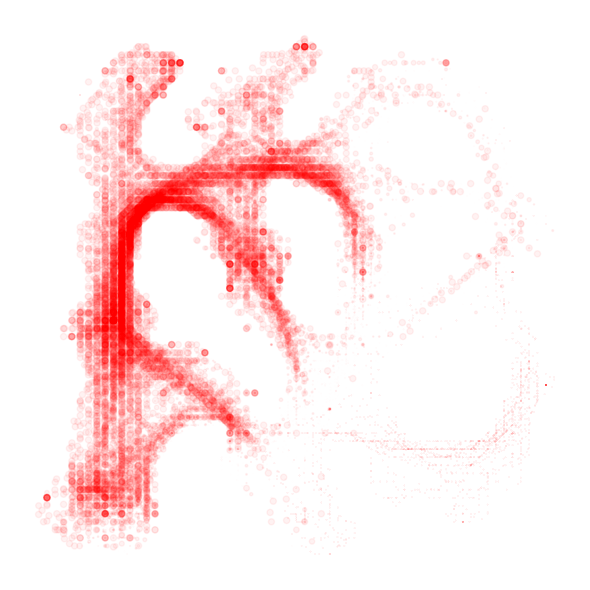}
	\label{fig_atttrajs}}\hspace{.05\linewidth}
	\subfloat[Features $f$]{\includegraphics[width=.4\linewidth]{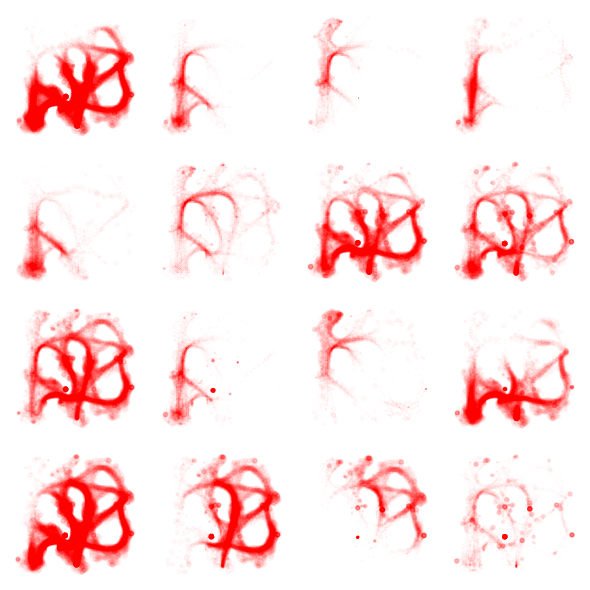}
	\label{fig_feattrajs}}\\
	\subfloat[$a \times f$]{\includegraphics[width=.4\linewidth]{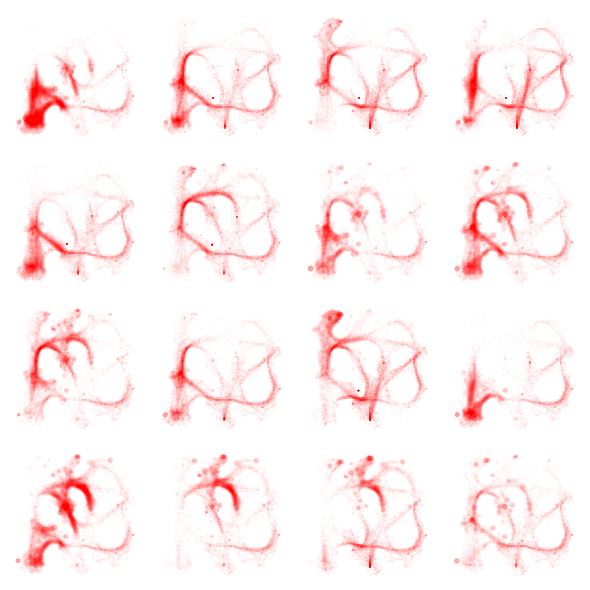}
	\label{fig_aftrajs}}\hspace{.05\linewidth}
	\subfloat[$a \times f$ for a single trajectory]{\includegraphics[width=.4\linewidth]{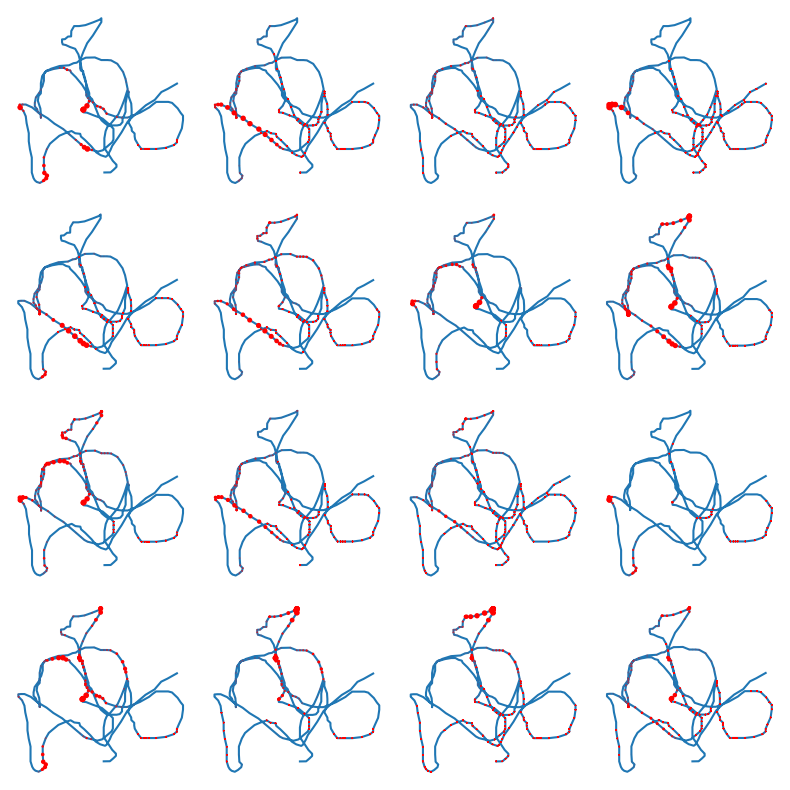}
	\label{fig_afsingletraj}}
	\caption{Plots of the output of the convolution. The size of the dot represents the activation of the output at this point. Transparency represents the density of activation at this point.}
	\label{fig_cnnplots}
\end{figure}

As it is the most visual, we will here explore the weights of the CNN component of the CompSNN (see \figurename~\ref{fig_cnnplots}). It has two components, a convolutional 10-to-16 features layer, and a 10-to-1 attention layer. We can visualize both those outputs, or the $feature \times attention$ (noted $f \times a$) output. \figurename~\ref{fig_atttrajs} shows a visualisation of where the attention layer activates, and \figurename~\ref{fig_feattrajs} shows how and where each feature from the feature layer is activated. Finally, \figurename~\ref{fig_aftrajs} shows the output of the convolutional part of the CNN. As we can see, while the attention layer's output seems to be spatially correlated, not all features \textit{look} at the same locations, and they seem to identify specific behaviours, as we can see in \figurename~\ref{fig_afsingletraj}, where some turns and straight lines are identified to be important by the model.

\section{Conclusion}
in this paper, we proposed a system able to capture spatio-temporal markers of trajectories followed by navigators with specific demographic information. By combining simple explainable models capturing different dimension of the signal, we were able to build an explainable model performing significantly better than its modules taken in isolation.
Future research could investigate the explainability of the GCNN module, implementing it using Koenecker delta localization, and enriching the model of other signal representations to make it more performant and give more insight on how the behavior relates to the demographics. It would also be interesting to study how the definition of the graph and of its signal influence the performances and explainability of the model. Beyond spatial navigation trajectories, this method could also be applied to other types of time series, such as eye-tracking datasets or other trajectory-like signals.

\section*{Acknowledgment}

This project was partially funded by the \textit{RFI Atlantic 2020} and \textit{RFI Ouest Industrie Creative} programs of the French region Pays de la Loire

\bibliographystyle{IEEEtran}
\bibliography{thesis-hdubois-2019}
\vspace{12pt}

\end{document}